%% file: root.tex
\documentclass[letterpaper, 10 pt, conference]{ieeeconf}  

\IEEEoverridecommandlockouts                              

\usepackage{graphicx,import} 
\usepackage{transparent}
\usepackage{amsmath} 
\usepackage{amssymb}  
\usepackage{commath}
\usepackage{paralist}
\usepackage{algorithm}
\usepackage{algpseudocode}
\algnewcommand{\IfThenElse}[3]{
  \State \algorithmicif\ #1\ \algorithmicthen\ #2\ \algorithmicelse\ #3}
\algnewcommand\algorithmicforeach{\textbf{for each}}
\algdef{S}[FOR]{ForEach}[1]{\algorithmicforeach\ #1\ \algorithmicdo}
\usepackage{lipsum}  
\usepackage{subfigure}
\usepackage{epstopdf}
\usepackage{filecontents}
\usepackage[misc]{ifsym}
\usepackage{color}
\usepackage{xcolor}
\usepackage{xxcolor}
\usepackage{pgf}
\usepackage{pgfplots}
\usepackage{tikz}
\usepackage{setspace}
\usepackage[skins]{tcolorbox}
\usepackage{todonotes}
\usepackage{multirow}
\usepackage{array}
\usepackage{svg}
\usepackage{footmisc}
\usepackage{url}
\usepgfplotslibrary{groupplots,fillbetween,}
\usetikzlibrary{arrows,shadows,petri,automata,shapes.geometric,patterns}
\pgfplotsset{compat=newest}

\title{\LARGE \bf
 Swarm Relays: Distributed Self-Healing Ground-and-Air Connectivity Chains
}

\author{Vivek Shankar Varadharajan, David St-Onge, Bram Adams and Giovanni Beltrame
  \thanks{Mr. Varadharajan, Dr. Adams and Dr. Beltrame are with the Computer and Software Engineering department, Polytechnique Montreal, e-mail: g.beltrame@polymtl.ca, Dr. St-Onge is with the Department of Mechanical Engineering, \'Ecole de Technologie Sup\'erieure, Montreal.
    Supplementary material: \protect\url{https://mistlab.ca/papers/SwarmRelays}}%
}

\begin{document}
\maketitle
\thispagestyle{empty}
\pagestyle{empty}
\graphicspath{{figures/}}
\begin{abstract}
  The coordination of robot swarms -- large decentralized teams of
  robots -- generally relies on robust and efficient inter-robot
  communication. Maintaining communication between robots is
  particularly challenging in field deployments where robot motion,
  unstructured environments, limited computational resources, low
  bandwidth, and robot failures add to the complexity of the problem.
  In this paper we propose a novel lightweight algorithm that lets a
  heterogeneous group of robots navigate to a target in complex 3D
  environments while maintaining connectivity with a ground station by
  building a chain of robots. The fully decentralized algorithm is
  robust to robot failures, can heal broken communication links, and
  exploits heterogeneous swarms: when a target is unreachable by
  ground robots, the chain is extended with flying robots. We test the
  performance of our algorithm using up to 100 robots in a
  physics-based simulator with three mazes and several robot failure
  scenarios. We then validate the algorithm with physical platforms: 7
  wheeled robots and 6 flying ones, in homogeneous and heterogeneous
  scenarios in the lab and on the field.
\end{abstract}

\section{INTRODUCTION}
Swarm robotics is a field of engineering studying the use of large
groups of simple robots to perform complex
tasks~\cite{Brambilla2013}. Ideally, a single robot failure in a swarm
should not compromise the overall mission because of the inherent
redundancy of the swarm~\cite{Brambilla2013}.  With robustness and
scalability, robotic swarms are foreseen as cost effective solution
for spatially distributed tasks.

In many such applications, the ability of the swarm to coordinate
depends largely on its ability to communicate. A reliable
communication infrastructure allows the robots to exchange information
at any time. However, real deployments include many potential
sources of failures (environmental factors, mobility, wear and tear,
etc) that can break connectivity and compromise the mission. In this
work, we address complete robot failures caused by a robot's inability
to communicate or to be detected by its neighbors. In addition, most
realistic applications require the robots to remain connected with an
operator at a ground station, to either provide information (e.g. in a
disaster response scenario) or to receive new commands (e.g. planetary
exploration).

\begin{figure}
	\centering
	\def\svgwidth{\linewidth}
	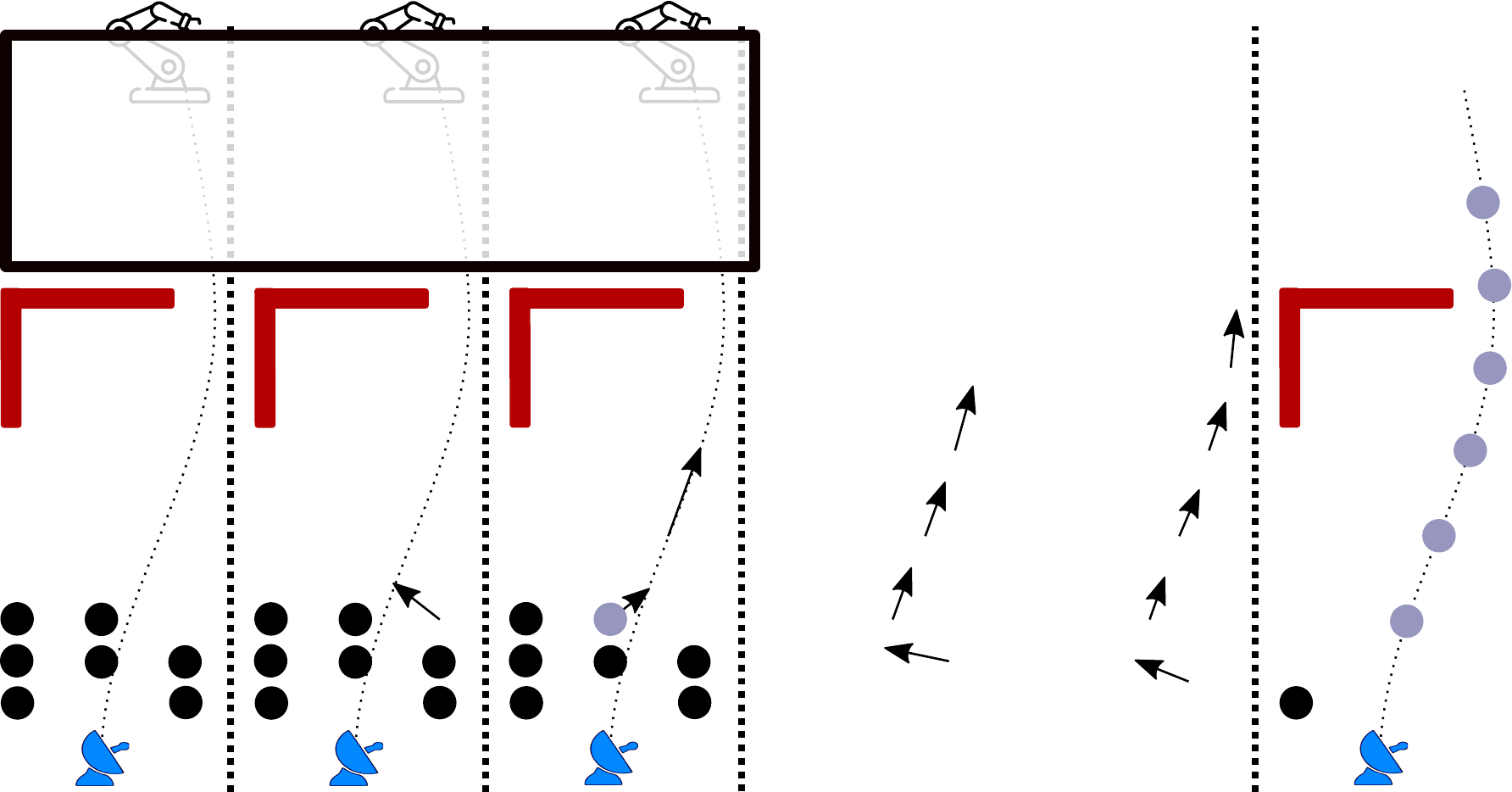
    \caption{Progressive formation of a chain to complete a task.}
	\label{fig:chain_formation_illustration}
\end{figure}

To maintain connectivity, we propose to progressively use the
available robots of a swarm to form a self-healing communication chain
from a ground station to a target (illustrated in
Fig.~\ref{fig:chain_formation_illustration}).  The operator sets the
target and the desired number of redundant links. We target
application scenarios including video relay and tele-operation for
exploration tasks, where a consistent end-to-end link is required. Our
algorithm uses a common path planner to generate a viable path, and
builds a chain of robots towards the target. This work
extends~\cite{varadharajan2019unbroken}, which briefly presented the
self healing aspects of this work without modeling and most
experiments. The contributions of this work are:
\begin{inparaenum}
\item a mathematical formulation of our self-healing chain formation
  algorithm with configurable redundancy;
\item the performance evaluation of our algorithm in simulated
  environments of various complexity and integrating robot failures;
\item the validation of the algorithm with a physical swarm of 7 wheeled
  robots and 6 flying robots, in homogeneous and heterogeneous configurations.
\end{inparaenum}

\section{Related work\label{sec:literature}}

There are two general approaches to connectivity maintenance in
multi-robot systems: strict end-to-end
connectivity~\cite{Stephan2017,majcherczyk2018decentralized},
or relaxed intermittent
connectivity~\cite{Kantaros2019,Guo2017,Hollinger2010}. While the
first demands to maintain a link from the source to the sinks (the
ends of each branch of the network graph), the latter allows for
momentary local breaks in the communication topology.
When the mission requires to continuously relay information, like
video, operators commands or offloaded computation, strict end-to-end
connectivity might be preferred; this is the approach we use in this work.
 
\noindent \textbf{Algebraic Connectivity}
The problem of preserving
end-to-end connectivity is widely discussed in recent
literature~\cite{Stephan2017,majcherczyk2018decentralized,panerati2018robust}.
Some works use continuous control models with algebraic
connectivity~\cite{sabattini2011decentralized} and implement
mission-related control laws~\cite{Siligardi2019}, considering robot
failures. De Gennaro et al.~\cite{de2006decentralized} derive a
gradient-based control law from Laplacian matrices' features such as
the Fiedler vector to maximize connectivity. Stump et
al.~\cite{Stump2008} also use Fiedler value and k-connectivity to
create a bridge between a station and a robot moving
towards a target in walled environments.
However, the centralized computation of the Laplacian is hardly
scalable, and distributed estimations require significant communication
bandwidth and are sensitive to noise~\cite{jacoicra2018}.

\noindent \textbf{Hybrid approaches} 
Connectivity can also be maintained by merging continuous motion
controllers and discrete optimization for packet
routing~\cite{Zavlanos2013}. INSPIRE~\cite{Williams2014} uses a two
layer control to preserve connectivity: the first layer applies a
potential field local controller to maintain a connected configuration
and the second layer optimizes the routing.  These methods have long
convergence time, and are generally not suitable for large robot
groups. Ji and Egerstedt~\cite{ji2007distributed} integrate connectivity
preservation in two control laws for rendezvous and pattern
formation. Their controller does not explicitly take into account
obstacles and hence cannot easily be adapted to cluttered
environments.

\noindent \textbf{Tree-based approaches}
Majcherczyk et al.~\cite{majcherczyk2018decentralized} deploy multiple robots
towards different targets while preserving connectivity in a decentralized
method based on tree construction. 
However, they carry all available robots along the path using a
virtual communication force field. This approach can lead to unwanted
redundant sub-structures and recurrent reconfiguration.  Hung et
al.~\cite{Hung2019} propose a decentralized global network integrity
preservation strategy that performs strategic edge addition and
removal to the network to maintain connectivity while reaching its
targets. This approach considers coverage missions and decomposes the
space into cells from which to select targets. However, distant 
and sparse targets increase convergence time significantly.

%
\noindent \textbf{Planner Based Approaches} Approaches leveraging a
common path planner (centralized~\cite{Fink2013} or
hybrid~\cite{Stephan2017}) determine the optimal communication points
for reliable connectivity. These works use a variant of
RRT~\cite{Karaman2011} that integrates a communication model to
estimate connectivity levels. These methods have realistic
communication models but they rely on a centralized solution (a
mission planner) and are computationally heavy (they solve a
second-order cone program -- SOCP).
Our approach is fully distributed: we estimate a path using an elected
robot whenever a new target becomes available, and navigate while preserving
connectivity (similar to~\cite{Zavlanos2007}).
Our method also allows for intermediate robot failures and changes in
the environment, which are not considered in the above solutions.

\noindent \textbf{Robustness to Failure} Connectivity maintenance for
multi-robot systems is a widely covered domain but very few works
address robot failures. Some approaches~\cite{panerati2018robust} take
into account a robustness factor to tackle robot failures but cannot
recover a completely disconnected network. A few other approaches
consider disconnection and recovery due to environmental
mismatches~\cite{marchukov2019multi} but do not consider complete
robot failures. The Wireless Actor Networks domain has several works
that address failure recovery~\cite{abbasi2007distributed}, e.g. using
dynamic programming~\cite{akkaya2009distributed} to reconnect a
disjoint graph. These approaches disregard motion planning and have
limited applicability in cluttered environments.

\noindent \textbf{Task allocation}
Allocation of a fixed number of
tasks to a set of robots is a well know combinatorial problem and
requires heuristics~\cite{pillac2013review} to approximate to a
polynomial solution. Decentralized methods use local planners along
with consensus algorithms to agree on the context of the task
plan~\cite{Shima2007}. Other approaches compute a plan on each robot
and use consensus algorithms to agree on the global
assignment~\cite{Choi2009}. We use an approach similar to the latter
to assign roles: a local bid on each robot serves to achieve consensus
on the assignments.

Our work leverages the ideas in~\cite{Hung2019}
and~\cite{majcherczyk2018decentralized} to dynamically build
structures towards the targets. We sequentially add edges to a tree in
a distributed manner by using the path from a standard path planner,
as in~\cite{Stephan2017}, and reactively enforce connectivity, as
in~\cite{Zavlanos2007}. The final contribution of our approach is the
ability to recover from simultaneous robot failures while navigating
complex environments with obstacles and preserving a network structure
with a configurable number of redundant links.
On top of this, our approach uses minimal computation and
communication load on the robots, a key aspect for the deployment of
other behaviors on top of connectivity maintenance. To the best of the
authors' knowledge, this is the first approach that studies this problem
in a holistic manner: a path planner, a failure recovery mechanism and
a task allocation mechanism to dynamically assign tasks.

\section{Preliminaries\label{sec:model}}
Consider a team of $N_r$ robots with their positions denoted by
$X=\{x_1,x_2, ..., x_n\}, \forall x_i \in \mathbb{R}^3$. The evolving position
of the robots at time $t_i$ can be denoted as $X(t_i) \in
\mathbb{R}^{3N}$. Given a set of target locations
$\mathbb{T} = \{\tau_1, \tau_2, ..., \tau_n\}, \forall \tau_i \in
\mathbb{R}^{3}$, our objective is to drive the robots to a formation that
ensures (a) at least one communication path between any two robots; and (b)
each target is within range of at least one robot
$\norm{x_w - \tau_i} \leq \delta_{tol} \forall \tau_i \in \mathbb{T}$. We
consider a single integrator robot model ($\dot{x}_i(t) = u_i$) and assume
that the robots are fully controllable with $u_i$. Taking into account that
the robots' workspace, $\mathbb{X} \in \mathbb{R}^3$, is divided into
obstacles $\mathbb{X}_{obs}$ and free space $\mathbb{X}_{free}$, we derive the
control inputs $u_i(t)$ for all robots, avoiding obstacles and other robots.

\subsection{Communication model}

We assume the robots are equipped with a wireless communication device
(e.g. 2GHz, 5GHz or 900MHz). The received signal strength is
influenced by three main factors: path-loss, shadowing, and
fading~\cite{mostofi2009characterization}.
We approximate signal strength as a generic function of distance.
 
Inter-agent communication in a group can be then modeled as a weighted
undirected graph $\mathcal{G} = (\nu,\epsilon,A)$, with the node set
$\nu=\{r_1,..r_N\}$ representing the robots, and the edge set
$\epsilon = \{ e_{ij} \vert i,j \in \nu, i \neq j \}$, representing
communication links. A common approach to working with a communication
graph is to use its adjacency matrix $A$, in which entries $e_{ij}$
represent the probability of robot $i$ decoding $j$'s packets. We aim
at maintaining $e_{ij} > e_{min}, \forall e_{ij} \in \epsilon$, that
is to guarantee a minimum signal quality.

We consider the robots capable of broadcasting and relaying
messages to their neighbors over a limited spherical communication
range $Z$. The robots estimate $e_{ij}$ for their neighbors using
local information, if $d_{ij} > Z$ then $e_{ij} = 0$, whereas if
$d_{ij} < \delta$ then $e_{ij} =1$ and $e^{\frac{-5*d_{ij}}{Z}}$
otherwise. $\delta$ is a small constant, slightly larger than the
radius of the robot. The approximation of connectivity using $e_{ij}$
allows for modeling additive white noise in sensing. The surroundings
of robot $i$ are divided into communication zones:
\begin{itemize}
\item the safe zone $Z_{i}^{s}$, in which neighbors are considered to
  have a reliable network link ($e_{ij} > e_{min}$) up to the limit
  distance $d_s$ with connectivity $e_{ij} = e_{min}$;
\item the critical zone $Z_{i}^{c}$, in which neighbors are
  getting close to the limit of the communication range. In this zone, $e_c$
  $\triangleq$ $d_c$ - $d_s > 0$ is defined as the \emph{critical
    tolerance}. At the critical distance $d_c$,  $e_{ij} < e_{min}$;
\item the break-away zone $Z_{i}^{b}$, in which neighbors are expected to
  break their network link. In this zone, $e_b \triangleq d_b - d_c > 0$ is
  defined as the \emph{break-away tolerance}. At the break-away distance
  $d_b$, $e_{ij}\approx 0$ and at Z, $e_{ij} = 0$ with $e_z \triangleq d_b - Z > 0$.
\end{itemize}

Let \textit{$N_{i}$} be the neighbor set of robot $i$, which is
divided into:
$\textit{N}_{i} = \textit{N}_{i}^{s} \cup \textit{N}_{i}^{c} \cup
\textit{N}_{i}^{b}$.  Robot $j$ is called a safe zone robot of robot
$i$ if $x_j \in Z_{i}^{s}$, with $x_j$ its position vector. The set of
all neighbors within the safe communication zone of robot $i$ is:
$\textit{N}_{i}^{s} = \{ j \vert x_j \in Z_{i}^{s}, \forall j \in \textit{N}_i \}$.
From this, we define the \emph{safe connectivity set} as the entries $e_{ij}$ of the
adjacency matrix $\forall j \in \textit{N}_{i}^{s}$. Similarly, we can define
the \emph{critical} and \emph{break-away} connectivity sets with the
corresponding entries of the adjacency matrix.
These sets allow us to derive control inputs that guarantee the
preservation of local connectivity.

\subsection{Local connectivity preservation}
We formulate the constraint for the preservation of connectivity among
the robots using geometric arguments as in~\cite{Hung2019}.
However, our solution continuously adds edges to the local graph until
specific robots with specific roles reach the targets. Our approach
reduces the graph construction time, computational load and
communication rounds required to determine which edges to remove
in~\cite{Hung2019}.

\begin{figure}
    \vspace*{0.3cm}
	\centering
	\resizebox{\linewidth}{!}{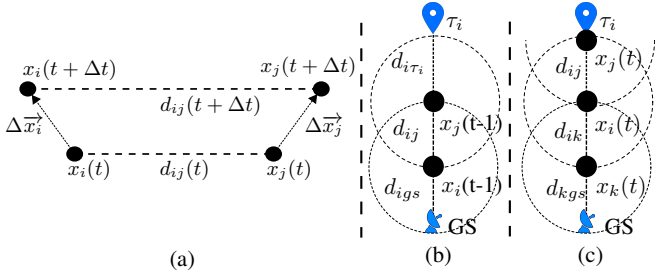}
	\caption{Variables related to inter-robot distance and robot
          position before and after a time step $\Delta$t (a)
          illustration of combined distance of a connectivity chain
          before (b) and after (c) the addition of a new robot.}
	\label{fig:movement}
\end{figure}
Fig.~\ref{fig:movement}(a) shows the initial position of robot $i$
($x_i(t)$) and robot $j$ ($x_{j}(t)$) with their relative distance
$d_{ij}(t)$, together with their new position and relative distance
after a time $\Delta t$. To guarantee the preservation of the
communication link, the control must ensure:
\begin{equation}\label{equ:theorem1}
	(\norm{\Delta x_i} + \norm{\Delta x_j}) \leq d_b-d_{ij} ,
\end{equation}
with $d_b$, the break-away distance where $e_{ij} = 0$, at which the
connectivity breaks. In our implementation we choose to split the
responsibility of respecting the available margin $d_b-d_{ij}$ equally
among the two robots.


\noindent\emph{Proof}: With $d_{ij}(t + \Delta t)$ the distance
between two robots $i$ and $j$ after a time step $\Delta t$ defined as
$d_{ij}(t + \Delta t) = \norm{x_i(t + \Delta t) - x_j(t + \Delta t)} =
d_{ij}(t) + \norm{\Delta x_i + \Delta x_j}$ . When we apply
(\ref{equ:theorem1}) we get:
\begin{equation}\label{equ:proof_need}
d_{ij}(t + \Delta t) \leq d_b.
\end{equation}
proving that robots $i$ and $j$ stay connected.    

We can extend the following remarks considering the three neighborhood sets
(safe, critical, break-away):
\begin{enumerate}
\item A robot $j$ can be in $\textit{N}_{i}^{s}$ if and only if, its
  position lies within the safe zone $Z_{i}^{s}$ of robot $i$. This
  implies that $d_b - d_{ij} \geq e_c + e_b$. If we choose control
  inputs that allow $\Delta x_i$ and $\Delta x_j$ to satisfy the
  condition of~(\ref{equ:theorem1}), with $d_s$ as the safe limit, the
  robots will always stay within the safe communication distance:
\begin{equation}\label{equ:proof_safe}
\Delta x_i \leq \frac{d_s - d_{ij}}{2} \quad and \quad 
\Delta x_j \leq \frac{d_s - d_{ij}}{2}  
\end{equation}
\item For robots $j \in \textit{N}_{i}^{c}$, located in the critical
  communication zone $Z_{i}^{c}$, if we choose control inputs that
  allow $\Delta x_i$ and $\Delta x_j$ to satisfy the condition
  of~(\ref{equ:theorem1}), with $d_s$ again as the safe limit, the
  robots tend to regain safe connectivity:
\begin{equation}\label{equ:proof_critical}
\Delta x_j \leq (d_s - d_{ij}) \quad and \quad \Delta x_i = 0
\end{equation}
\item We can apply an identical reasoning for robots
  $j \in \textit{N}_{i}^{b}$ with their position within and break-away
  communication zone $Z_{i}^{b}$, have one robot stationary and the
  other apply a control input $\Delta x_i$ to satisfy the condition
  of~(\ref{equ:theorem1}).
\end{enumerate}
By applying control inputs satisfying the condition of
Equ.~(\ref{equ:proof_safe}) robots in critical/break-away connectivity
move towards each other to regain safe connectivity. We use the
critical tolerance $e_c$ and break-away tolerance $e_b$ to account for
control errors when applying Equ.~\ref{equ:proof_critical}.

\subsection{Global Connectivity Chains}
Given a group of robots, we build a chain from a ground station to a
target location. To build the chain, we assign roles and relationships
to the robots. Robots in the chain are assigned parent-child
relationships, to manage the construction of the chain
towards a target while maintaining connectivity.

\noindent\emph{Definition 1}: A connectivity chain is a tree $C$ represented
as a partially ordered set $(C,<) = \{c_1,c_2,...,c_n\}$ with
$\vert C \vert = n$. All vertices $c_i \in C$, have a parent $c_{i-1}$ and a
child $c_{i+1}$, except for $c_1$, the \emph{root}, that is without parent and
$c_n$, the \emph{worker}, which is without children. All other $c_i$ are
\emph{networkers}. The edge set $\epsilon = \{e_{c_{i},c_{i+1}}\vert \forall c_i \in C\}$.

\noindent\emph{Proposition 1:} Given a connectivity chain $(C,<)$ and the
distance of the worker $c_n$ to a target $d_{i\tau_i}$, the addition of a new
robot between any two robots $c_i$ and $c_{i+1}$ in the chain decreases the
distance $d_{i\tau_i}$. Let $d(t-1)$ denote the sum of the distances between
all the robots in a chain before the addition of robot $c_k$, considering the
distances of the robots $d_{n-1,n}=d_s \forall n \in (C,<)$. The distance
$d$(t) after the addition of $c_k$ into chain is $d(t)=d(t-1)+d_s$,
adding a slack of $d_s$ into the chain, which in turn allows the worker to
move towards the target, decreasing $d_{i\tau_i}$, as shown in
Fig.~\ref{fig:movement}(b-c). 
The desired network structure from the root robot to a worker robot is
specified as the minimum number of mandatory communication links $C_n$
and the algorithm enforces $C_n$ individual connectivity chains.

\subsection{Task allocation problem}
Let the set of free robots $N_f = \{N_r\}/N_c$, with $N_r$ the set of
all the available robots and $N_c$ the set of robots in a connectivity
chain. The goal of the task allocation algorithm is to assign the set
of tasks $\mathbb{T}$ to robots in $N_f$. The problem of assigning
each robot to a single task is commonly refereed to as Single
Assignment (SA) problem~\cite{Choi2009}. The resulting task allocation
problem takes the following form:

\begin{subequations}
\begin{alignat*}{2}
&\!\min        & & \sum_{i=1}^{|N_f|} \sum_{j=1}^{|\mathbb{T}|} c_{ij}(x_i)a_{ij}\\
&\text{subject to} &      & \sum_{j=1}^{|\mathbb{T}|} a_{ij} \leq 1, \forall i \in  N_f \text{ and }  \sum_{i=1}^{|N_f|} a_{ij} \leq 1, \forall j \in \mathbb{T} 
\end{alignat*}
\end{subequations}

\noindent where $c_{ij}(x_i) = || x_i - \tau_{j} ||$ denotes the cost
of robot i at position $x_i$ performing task j and
$a_{ij} \in \{0,1\}$ is the assignment variable indicating the
assignment of task j to robot i. The first constraint indicates that
each robot can be assigned at most one task and second indicating no
two robots get the same task. Consensus-based auction algorithm
(CBAA)~\cite{Choi2009} uses an auction to obtain initial bids from
robots and unifies the assignment using a consensus phase. CBAA
guarantees convergence and provides a bound on optimality when the
cost function follows a diminishing marginal gain~\cite{Choi2009}. In
this work, we use Virtual Sigmergy~\cite{Pinciroli:vs:2016} to share
task assignments in a decentralized manner. Virtual Stigmergy is a
decentralized information sharing mechanism: once a value is written
in the Virtual Stigmergy by a robot, it can be accessed from all other
robots.
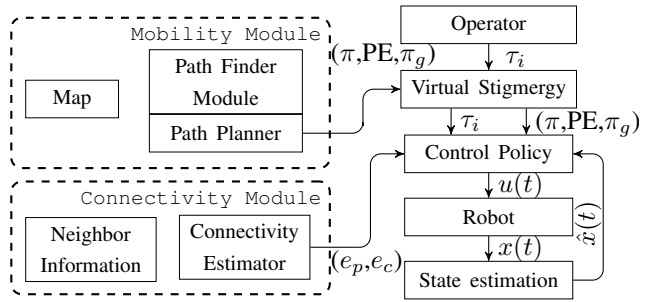
\begin{figure}[tbp]
	\centering
	\input{figures/method.tex}
	\caption{Control architecture used within the chain construction algorithm.}
	\label{fig:method} 
\end{figure}%
\section{Proposed Method \label{sec:approach}}
Fig.~\ref{fig:method} shows the modules of our controller: mobility,
connectivity and control policy. The mobility module computes a viable
path optimized for path length and stores the result in the Virtual
Stigmergy. The connectivity module continuously estimates the current
connectivity between the assigned parent ($e_p$) and child
($e_c$). The control policy computes the control inputs using
path and connectivity information.

\label{subsec:path_planner}
We consider two types of robots in the swarm:
\emph{ground robots} with state space $\mathbb{X} \in \mathbb{R}^2$,
and \emph{flying robots} with $\mathbb{X} \in \mathbb{R}^3$. We assume the map of the environment is
known.

The mobility module consists of a path finder and an optimal path planner. The
path finder scans for a viable path by splitting the workspace in cells at
a given resolution $R$, and building a graph $\mathcal{G_{PE}} = (N,E)$ of the
cells. It then performs a depth-first-search in this graph to
verify the reachability of target $\tau_{i}$~\cite{Zhang2006}. If it is found reachable, the module raises the \emph{PE} flag. Ground robots and flying robots
can then be used to build a chain on the resulting path $\pi$.
If a path to $\tau_{i}$ is not found, it either means that the target is
unreachable or that the check was done in $\mathbb{R}^2$, but a solution
exists in $\mathbb{R}^3$. The path finder can be set to always check in
$\mathbb{R}^3$, but that would create overhead for simple cases in
$\mathbb{R}^2$. Nevertheless, the resulting path in $\mathbb{R}^3$ has a
portion, $\pi_g \subset \pi$, that can be traveled by ground robots: the projection of the 3D path $\pi$ on the ground
plane.

The path planner module computes an optimized path (continuous vector-valued
function) $\sigma : [0,T] \to \mathbb{X}_{free}$ such that $\sigma(0) = x_{0}$
and $\sigma(T) = \tau_{i}$. Similar path planning problems have led to several
well-studied approaches~\cite{Karaman2011,Li2015}.  When optimizing for the
shortest path length, sample-based approaches can quickly get an
approximation that is then optimized.
We selected $RRT^*$ because it was shown to be faster in large environments
and to perform well in cluttered spaces~\cite{Karaman2011}. However, our
control architecture is agnostic to the path planner algorithm, and we have
tested it with several other planners (SST, $BIT^*$ and $PRM^*$) with comparable
results, as discussed in~\ref{sec:experiments}.

The chain construction algorithm is implemented in
Buzz~\cite{PinciroliBuzz2016}, a programming language for robot swarms, which
eases behavior design efforts by providing several swarm programming
primitives. For the planners, we integrated classes of Open Motion Planning
Library (OMPL)~\cite{sucan2012open} in our architecture to be accessible as
Buzz functions.
\begin{figure}
	\centering
	\input{figures/interaction.tex}
	\caption{Interplay between Networkers and Workers.}
	\label{fig:interplay} 
\end{figure}
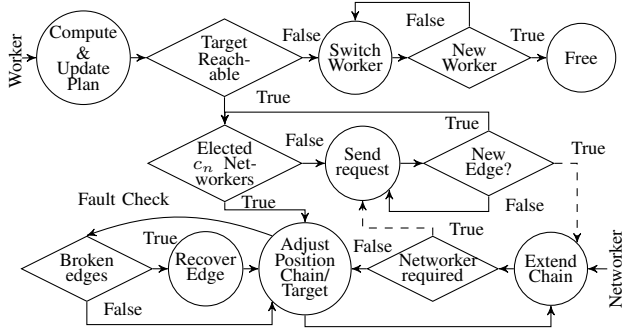%

We assigned one of three roles to the robots: root, worker(s), and
networkers. A detailed presentation of the task allocation
strategy is given as supplementary material.

At launch, the root is either pre-determined (ground station) or
elected using gradient algorithm (detailed in the supplementary
material). Each target also gets a
worker robot elected with the same strategy. The gradient algorithm is
implemented using Virtual Stigmergy~\cite{Pinciroli:vs:2016}.

Fig.~\ref{fig:interplay} shows the interaction of the workers and
networkers in the chain formation. The worker robot checks for the
existence of a path tuple $\langle\pi,PE,\pi_g\rangle$ in the Virtual
Stigmergy; if not available, it computes the path and shares it. The
motivation behind using a single robot to compute the plan is
twofold: \begin{inparaenum}\item if multiple robots are computing a
  plan then all plans must be exchanged to reach consensus -- a high
  load of messages/bandwidth; \item assigning the task to the worker
  robot also leads to more efficient updates, as the worker will be
  the first to encounter any obstacles in the
  environment\end{inparaenum}. In case of a mismatch between the
computed path and the explored environment, the path can be adapted.

\begin{figure}[tbp]
\centering
\resizebox{1.1\linewidth}{!}{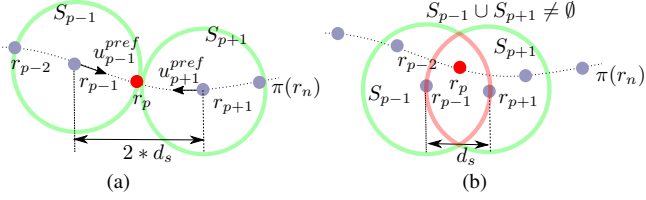}
\caption{Failure recovery: (a) at the time of failure and (b) after recovery.}
\label{fig:fault_illus} 
\end{figure}
%

%
The path's tuple includes $\pi$, the path defined as a sequence of
points (states), the \emph{PE} flag and $\pi_g$, the projection of
$\pi$ on the ground plane. Following its locomotion type, the robot
knows which sections of $\pi$ it can reach. If the elected worker
robot does not have the required locomotion type, the swarm elects a
more suitable robot, as shown in Fig.~\ref{fig:interplay}. The worker
robot determines the number of links $C_n$ required to reach the
target and elects the necessary networkers. If more edges are required
than the number of robots surrounding the worker, the request is
passed down the networkers towards the root, until a free robot
(future edge) is found. An optional module, the Way-Point (WP)
prediction, grants each robot in the chain to autonomously elect
robots to expand the chain without a signal from their child. The
number of robots required in a chain $n = ceil( \frac{d_{path}}{d_s})$
is estimated on each robot using the current plan length $d_{path}$.


\subsection{Motion Control}
The motion commands are computed by networkers and workers (the root is fixed)
using the available path to extend the chain towards the target(s).
The robots that do not belong to any chain are called free robots, and
they wait close to the root until they get elected as a worker or a
networker. The robots compute the preferred
velocity as:
\begin{equation}
u_{i}^{pref}=
\begin{cases}
f_c(d_{i}^{C}) u_{i}^{path}(\pi,F) &\text{if } d_{i}^{P} \leq d_s\\
u_{i}^{path}(\pi,B) &\text{otherwise}
\end{cases}
\end{equation}
\begin{equation}
f_c(d_{i}^{C})=
\begin{cases}
1 &\text{if } d_{i}^{C} \leq d_s\\
0 &\text{otherwise}
\end{cases}
\end{equation}
\noindent where $u_{i}^{path}(\pi,F)$ computes velocity commands to move the robot
towards the target and $u_{i}^{path}(\pi,B)$ computes velocity commands to
move the robot towards the root, both based on the path
$\pi$. $f_c(d_{i}^{C})$ stops the movement when a child reached the critical
communication distance from its parent. In other words, a chain retracts if a robot is in
the critical communication zone and expands otherwise.
The inputs $u_{i}^{pref}$ are computed from set point control and altered by a
collision avoidance algorithm. Most reactive decentralized collision avoidance
can be used within our architecture. We used the Reciprocal Velocity
Obstacle (RVO) algorithm~\cite{van2008reciprocal}, which selects the best
velocity considering all future potential collisions. The resulting control
law is:
\begin{equation}\label{equ:rvo}
u_{i}=\text{arg }\min_{u_{i}^{\prime}\in Au_{i}} \sum_{j\in N_{i}^{close}} \alpha \frac{1}{tvo_{j}(u_{i}^{\prime})} + \norm{ u_{i}^{pref} - u_{i}^{\prime} }
\end{equation}
\begin{equation*}
Au_{i}=\{u_i^{\prime} \ \lvert \ \norm{u_i^{\prime}} < u_{i}^{conn} \}, \ u_{i}^{conn}=\min_{j \in P \cup C} \frac{(d_s-d_{ij})}{2\Delta t}
\end{equation*}
Equation~\ref{equ:rvo} computes the control input from the velocity within the
admissible set $Au_{i}$ that minimize the risk of collision while maximizing
the fit to the path. The function $tvo_{j}(u_{i}^{\prime})$ computes the
penalty of future collision between $i$ and $j$ while the norm is a penalty
for deviation from the preferred velocity on the path $u_{i}^{pref}$. $\alpha$
is a scaling factor to balance the penalty terms (set to 1 in our
tests).
$N_{i}^{close} =\{ j \ \lvert \ d_{ij} < r_{col} \forall \ j \ \in N_{i}\}$ is
the set of neighbors close enough to be inside the collision radius $r_{col}$.
$u_{i}^{conn}$ is the maximum velocity allowed, a bound to maintain a safe
communication distance according to Equation~\ref{equ:theorem1}.
%
%
%
\begin{figure}
\vspace*{0.15cm}
\centering
\includegraphics[width=0.99\linewidth]{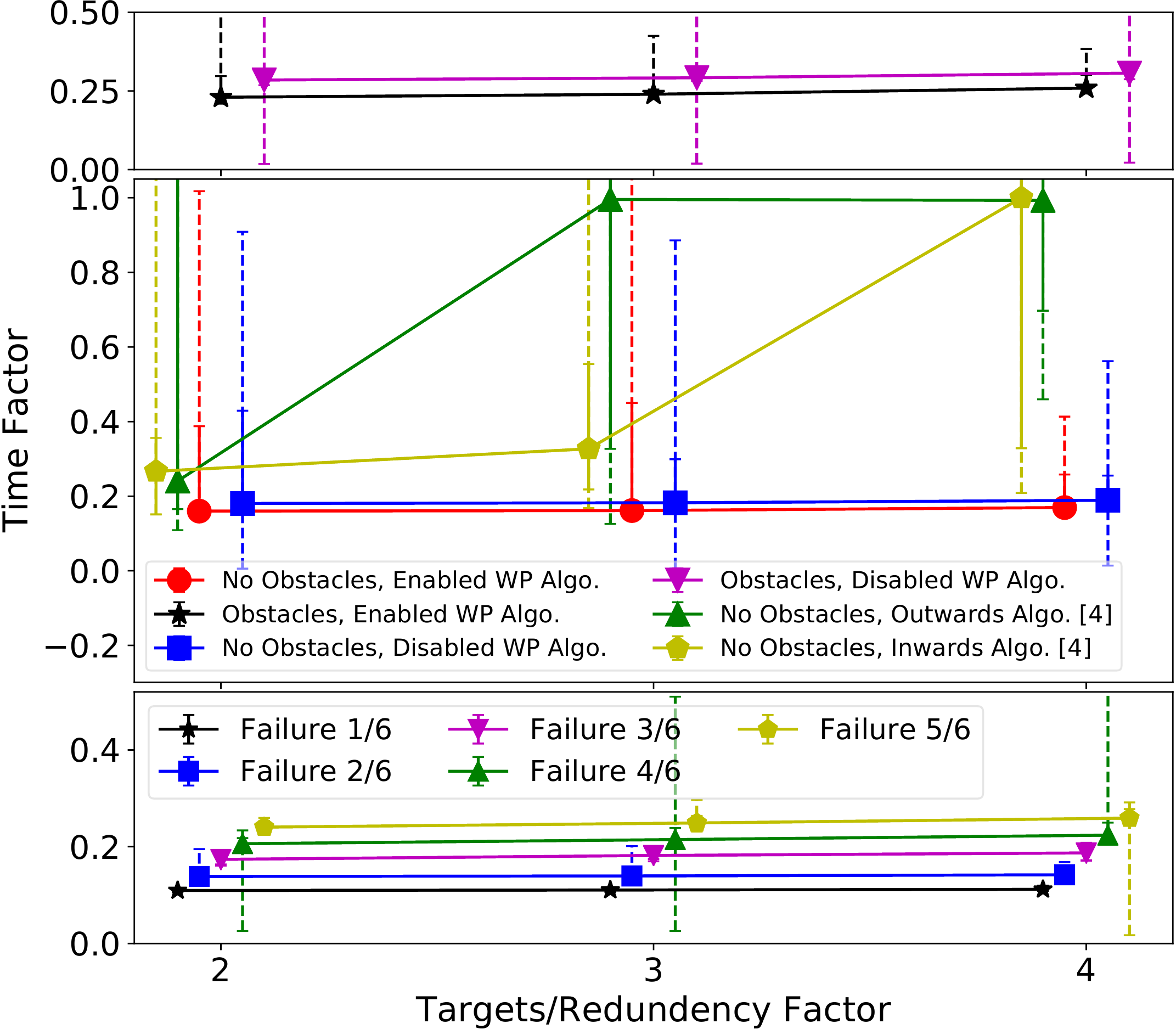}
\caption{Middle: comparison of convergence time with~\cite{majcherczyk2018decentralized},
  convergence time of the chain construction algorithm by enabling and
  disabling the way-point algorithm that allows the robots in a chain to
  predict the number of robots required. Top: convergence time by introducing one "C" shaped obstacle. Bottom: comparison of failure recovery time by failing different
  number of consecutive robots in a chain and creating two subgraphs.}
\label{fig:time_comp}
\end{figure}%
\subsection{Self-Healing}
We consider two types of robot failures in our work: due to sensing
errors and complete robot failures. Both cases create two or more
disconnected groups of robots that then need to be
reconnected. Dealing with complete failures is harder than 
sensing errors since no information transfer is possible between each
side of the disconnection.
Without any means of communication, it is impossible to agree on the
reconnection strategy between chain segments. In this work, we get the
information required to reconnect the chain in both cases using a periodic
broadcast message (detailed in the the supplementary
material).

Every robot $r_i$ in a chain $C$ uses a common plan $\pi(r_n)$ updated
by worker $r_n$ and shared through Virtual Stigmergy. They are aware
of $k$ local edges of their chain $C$ and its current depth through
the chain link messages (a sequence of IDs obtained through
broadcasts). Every robot can reconstruct the $k*2$ immediate portion
of the chain $C_{local} \subseteq C$ by concatenating parent and child
chain messages. If a robot in the communication path fails, the
communication chain is broken and the chain cannot be completed. In
this case, the parent and the child of the failed robot attempt to
bridge the broken link by navigating toward each other using the
available chain information as illustrated in the
Fig.~\ref{fig:fault_illus}.
\begin{figure}
\centering
\includegraphics[width=0.99\linewidth ,scale=0.99]{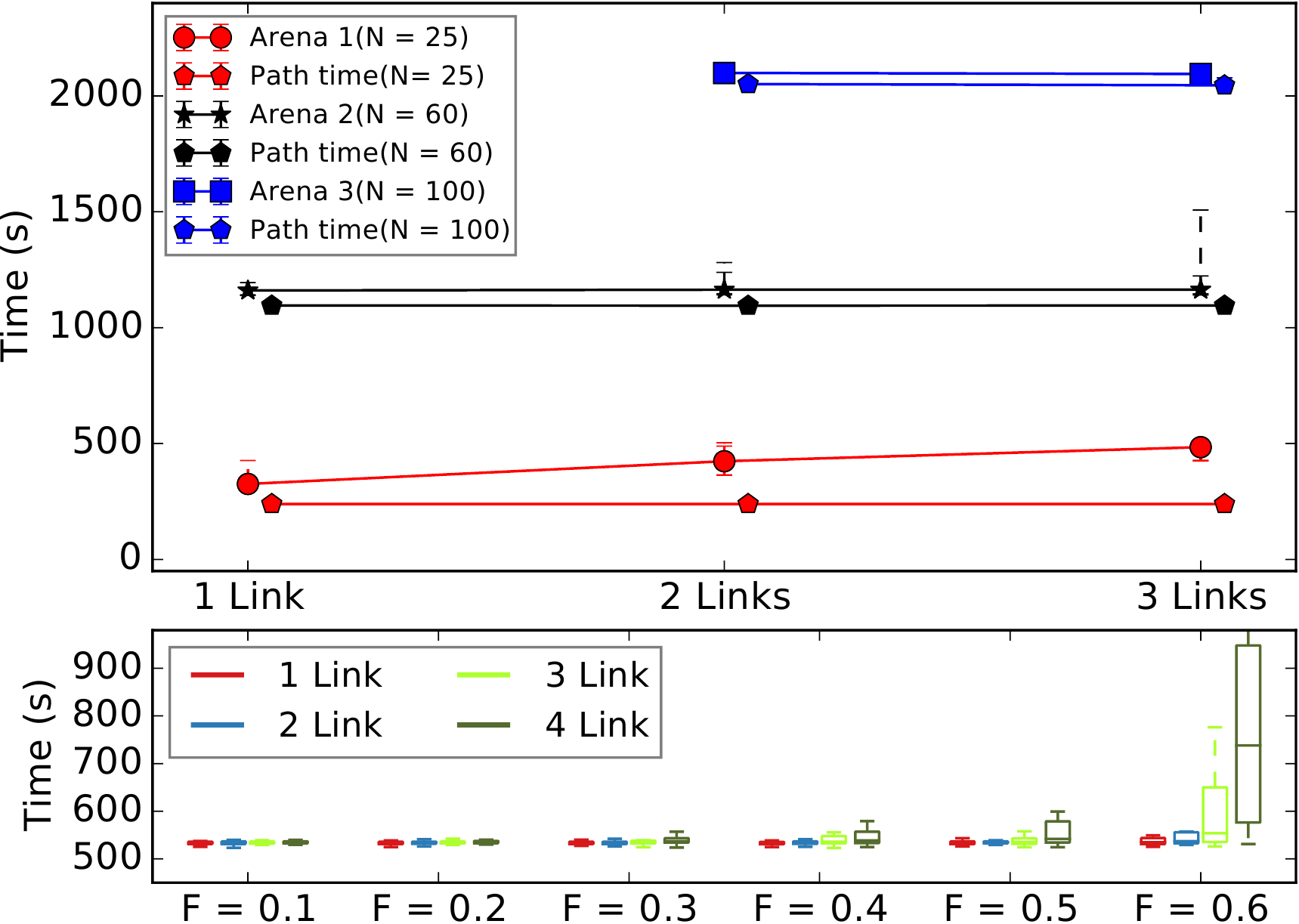}
\caption{Top: time taken by 25 (arena 1), 60 (arena 2) and 100 (arena
  3) robots to reach the target with 1, 2, and, 3 communication
  chains. The traveling time on the bottom indicates how fast a robot
  can travel the path without considering the chain. Bottom: time
  taken to build the communication with different percentage of random
  robot failures.}
\label{fig:sim_time}
\end{figure}%
Without loss of generality, we consider an arbitrary robot $r_p$ in a
chain failing with a parent $r_{p-1}$ and child $r_{p+1}$. The result
is two disconnected chains $C_1 = \{r_1,...,r_{p-1}\}$ and
$C_2=\{r_{p+1},...,r_{n}\}$. The robots $r_{p-1}$ and $r_{p+1}$ will
attempt to bridge the gap. The safe connectivity workspace
$S_{i} = \{y \in \mathbb{R}^n | \|y - x_i\| \leq d_s\}$ of robot $i$
contains all the states within the safe communication distance
$d_s$. At the time of failure $\{x_{p-1}\} \cap S_{p+1} = \emptyset$
and $\{x_{p+1}\} \cap S_{p-1} = \emptyset$, formally indicating the
disconnection. Robots $r_{p-1}$ and $r_{p+1}$ already share consensus
on the plan $\pi(r_n)$, when the robot is executing the current
segment of the plan. This allows the robots $p-1$ and $p+1$ to compute
control inputs $u_{p+1}^{pref} = u_{p+1}^{path}( \pi(r_n), B)$ and
$u_{p-1}^{pref} = u_{p-1}^{path}( \pi(r_n), F)$ respectively. When
using an identical plan $\pi(r_n)$ and applying $u_{i}^{path}$, it is
guarantied that the robots will regain safe communication distance
($d_{ij} \leq d_{s}$) with $x_{i} \in S_{j}$. The same reasoning
applies to $n$-consecutive robot failures, with the inter-chain
distance between two disconnected chains being $(n+1)*d_s$ instead of
$2*d_s$. The two bordering robots in the disconnected chain will
compute and apply the control input $u_{i}^{pref}$ based on
$\pi(r_n)$. The boundary robot that lost a child's connection acts as
a temporary worker until the chain is healed or it becomes the worker.


%
%

%
\section{Experimental Evaluation}
\label{sec:experiments}



\noindent\textbf{Performance comparison}
We use ARGoS3~\cite{Pinciroli:SI2012} physics-based simulator to
assess the performance of our method.  We compare our results
with~\cite{majcherczyk2018decentralized}, the closest approach to
ours.
We set up a simple open environment to compare our algorithm
with~\cite{majcherczyk2018decentralized} in 12 conditions: 2 to 4 targets
uniformly distributed on a circle ($r=10 m$), with and without obstacles
(C-shaped) between the robots and the targets, and with or without the WP
prediction algorithm mentioned in Section~\ref{subsec:path_planner}. Each
conditions was repeated 35 times with random initialization.

We set the algorithm's parameters to: $d_s = 1.4m$, $d_c = 1.6m$,
$d_b = 1.8m$, $v_{max} = 50 cm/s$, $\alpha=1$ and $r_{col}=25cm$, with a fixed
planning time of 2s, a bidding time for the gradient algorithm of 10s, a
forgetting time for status messages to 3s and a link failure declaration time
of 5s.


Fig.~\ref{fig:time_comp} shows the resulting comparison of performance. The
metric is \emph{time factor}, i.e. the time taken by the algorithm divided by
the traversal time~\cite{majcherczyk2018decentralized}. Our method outperforms
both the inwards and outwards algorithms proposed
in~\cite{majcherczyk2018decentralized}, particularly when increasing the
number of targets or redundancy factor. This can be attributed to the fact that
all robots are part of the large virtual force field
in~\cite{majcherczyk2018decentralized}, whereas we form a directed chain with
a near-optimal number of robots.
 

Unlike~\cite{majcherczyk2018decentralized}, our chain construction scales well
with the number of targets, showing the ability to reach separate targets in
parallel.  The time factor decreases by 2\% without obstacles and 5\% with
obstacles when using WP
prediction. 

\noindent\textbf{Robustness}
We then run the same configuration as above, but inducing up to 80\%
consecutive robot failures. Fig.~\ref{fig:time_comp} (bottom) shows
the time factor required by the robots to recover from these broken
links. There were in average 6 intermediate networker robots
connecting the root and the worker. The failure factor denotes the
number of consecutive robots disabled after the chain reached the
targets, for instance a factor of $5/6$ denotes failing 5 out of 6
robots in a row. Experiments show an increase in 5\% of the time ratio
for every additional robot failing in the chain. Simultaneously
failing a constant fraction of robots leads to almost constant
recovery time, as reconnections occur in parallel. The supplementary
material provides additional experiments that show
our method's robustness to noise.

To further assess the robustness of our method, we used a motion
planning benchmarking dataset~\cite{sturtevant2012benchmarks}:
\begin{inparaenum}
\item small (dragon age/arena),
\item medium (dragon age/den203d), and,
\item large (dragon age/arena2).
\end{inparaenum}, using the same parameters as above (except for
$d_s = 9.5m$, $d_c = 9.7m$, $d_b = 10m$ to fit the environment) but
injecting random robot failures at every control step with an
occurrence probably of $p=0.0005$ and bounded to a maximum percentage
of robots failures $F$ from the set
$F \in \{0.1,0.2,0.3,0.4,0.5,0.6\}$, with $ 0 \leq F \leq
1$. Fig.~\ref{fig:sim_time} (bottom) shows the resulting time taken to
build the chain. $F \leq 0.3$ does not have significant impact on the
algorithm performance. However, with more failures, completion time
increases as well as the performance variability. As more robots fail,
the links have to be repaired before proceeding to the target,
increasing the convergence time with $F$.
%
%
%
%
\begin{figure*}[tbp]
\vspace{0.15cm}
\centering
\includegraphics[width=0.99\linewidth ,scale=0.95]{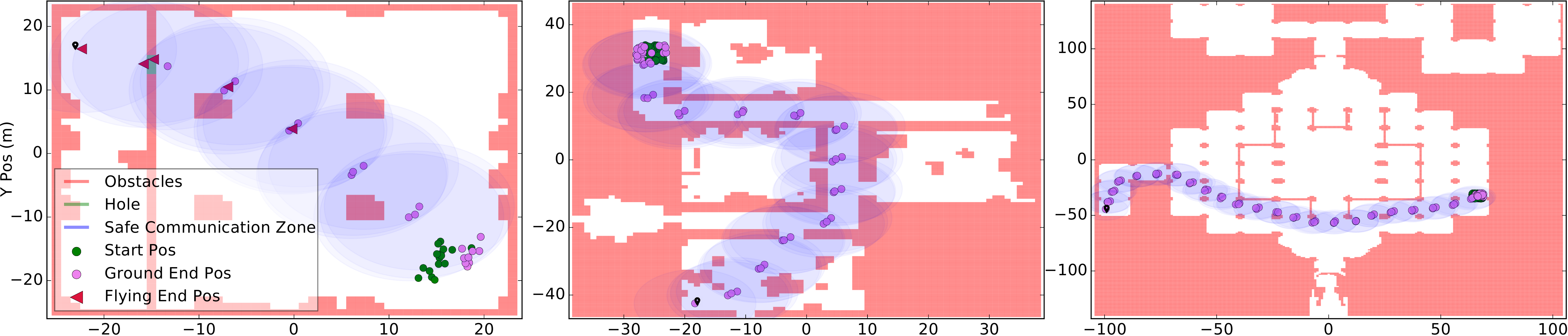}
\includegraphics[width=0.329\linewidth]{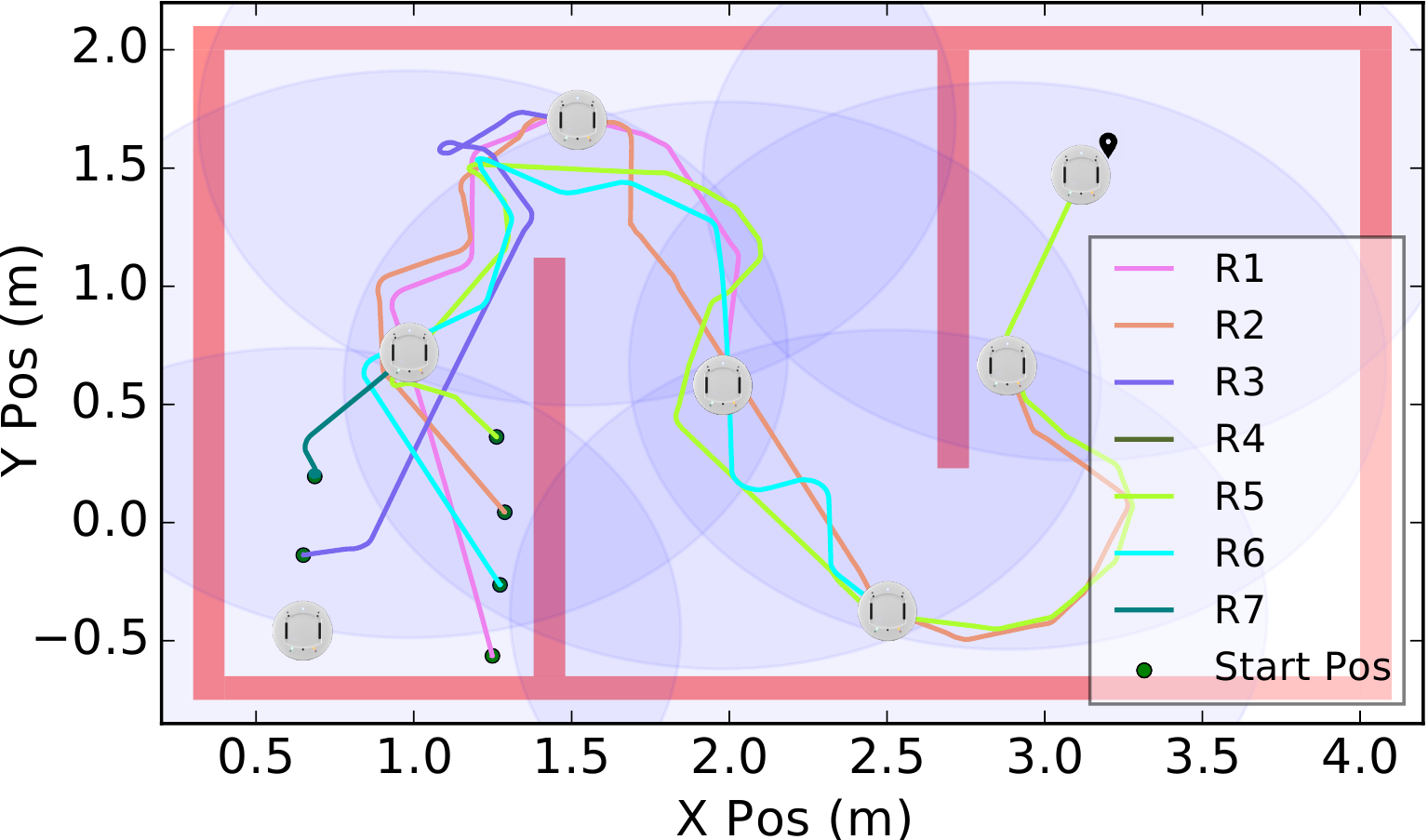}
\includegraphics[width=0.329\linewidth]{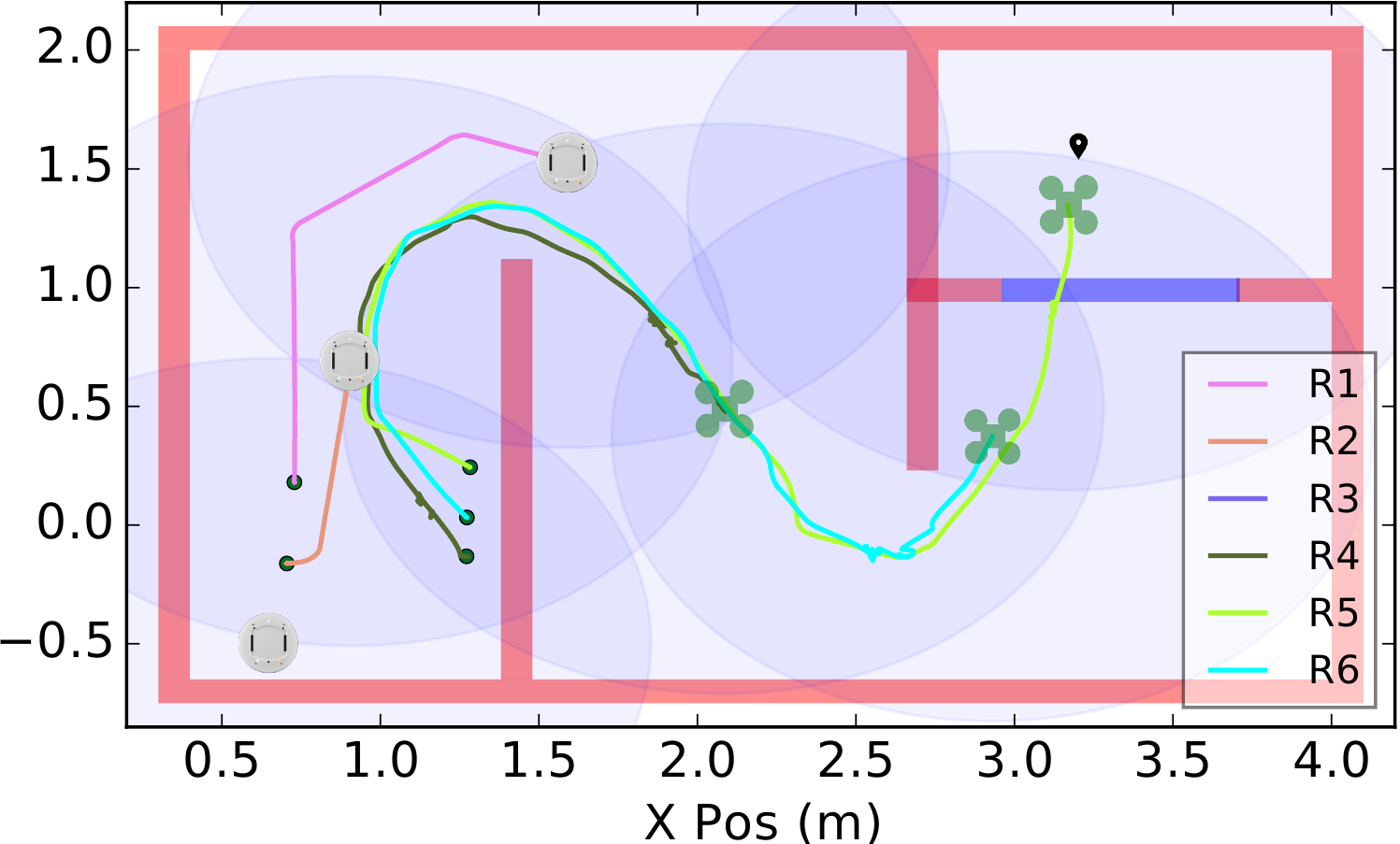}
\includegraphics[width=0.329\linewidth]{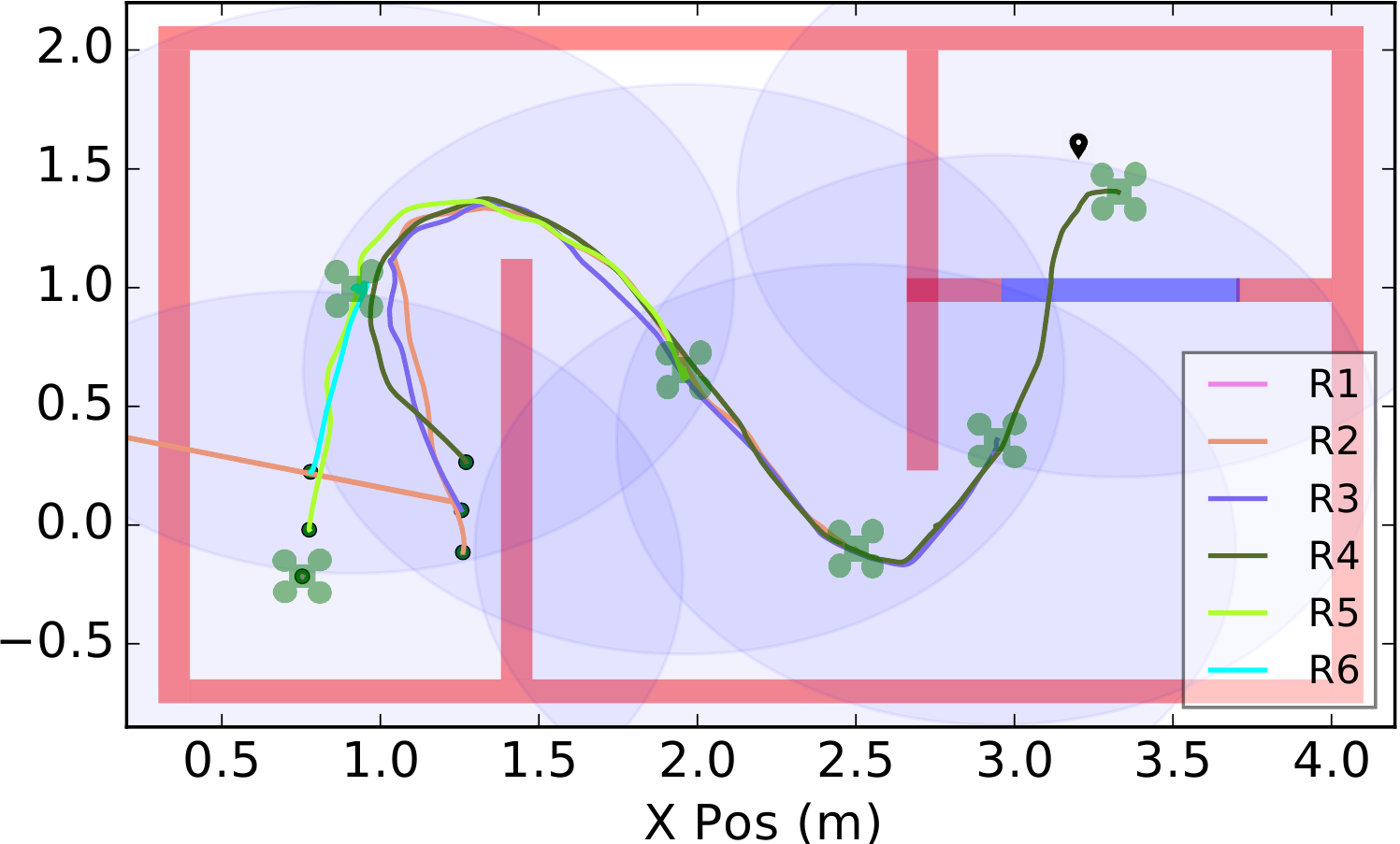}
\caption{Top: Illustration of simulated maps with each robot starting
  position, end position and the safe communication disk region: on the left
  the smallest arena with flying robots going over an obstructed zone, in the
  middle a medium-sized arena with ground robots only, and on the right our
  largest arena, also using only ground robots. Bottom: Real robot
  experimental arena with start, end, trajectory and end position: on
  the left 7 khepera IV robots in arena1, in the middle 3 khepera IV robots
  and 3 CrazyFlie in arena 2 and on the right 6 CrazyFlie in arena 2. }
\label{fig:sim_arena}
\end{figure*}%
%

%
%
%
%

\noindent\textbf{Scalability}
We use again the benchmarking dataset~\cite{sturtevant2012benchmarks},
but adding a wall with a small window above the ground to the first
arena to force the use of flying robots (top leftmost map of
Fig.~\ref{fig:sim_arena}). We used 20 ground robots and 5 flying
robots for arena 1, 60 ground robots for arena 2, and 100 ground
robots for arena 3. For each of the arenas, we enforced 1, 2, and, 3
communication links, except for the 100 robot runs (arena 3) covering
only 2 and 3 links. Fig~\ref{fig:sim_arena} (top) shows that the
robots stayed within the safe communication zone even when navigating
in narrow corridors. Fig.~\ref{fig:sim_time} (top) plots the time
taken by the robots to build 1, 2, and, 3 links in each of the arenas,
along with the reference traveling times computed using the path
length and the maximum velocity of the robots (set again to 10
cm/s). As expected, the time taken to build a chain increases with the
size of the arena, but the average time to build different number of
links is similar across different arenas. In particular, when
computing the time factor, the median time
ratio for N=60 is 0.106 and 0.102 for N=100, for all types of
links. This proves our claim that the proposed approach spends on
average equal amount of time reaching multiple targets with single or
multiple links. However, in arena 1 the chain construction time
increases with the number of links: this is due to the need of a
specific type of robot (flying) to cross a section of the arena. On
average, traversal time increases by 2\% for every additional
heterogeneous robot chain.


\noindent\textbf{Real robot experiments}
\label{subsec:robot_exp}
We validated the performance of our implementation using 6
Crazyflie
flying robots and 7
Khepera IV ground robots. We performed experiments
in three different settings with two different arena configurations as shown
in Fig.~\ref{fig:sim_arena}:
\begin{inparaenum}
\item 7 ground robots in arena 1,
\item 3 ground robots and 3 flying robots in arena 2;
  arena 2 has a wall containing a hole before the target and can be only
  reached by a flying robot,
\item 6 flying robots in arena 2).
\end{inparaenum}
The ground robots run an instance of the Buzz Virtual Machine (BVM)
on-board to perform both control and path planning, whereas the flying
robots use PyBuzz, a decentralized infrastructure detailed
in~\cite{cotsakisdecentralized2019}. We use a motion capture camera
system emulating situated communication for state estimation. For
these experiments, we set the design parameters to $d_s = 1.2m$,
$d_c = 1.4m$, $d_b = 1.5m$ and $v_{max} = 1.5 m/s$ to match the size
of our robots and arena.

Fig.~\ref{fig:sim_arena} (bottom) shows one of the 10 test runs we
performed in each of the three different settings. The figure shows the
starting point, the trajectory, and the end position of the robots. On all
runs the robots were able to build a communication chain and reach the target.
Ground robots, heterogeneous team and flying robots on an average took 100s, 50s and 40s respectively to reach the targets (as discussed in the supplementary material). 
Finally, we ported the algorithm to our ROS
based infrastructure~\cite{stonge2020} for an outdoor test on a group
of 4 larger quadcopters. The outdoor robots used GPS and communicated
through an Xbee mesh network. This deployment was made to show the
practical use of our algorithm on commercial hardware.

\section{Conclusions}
\label{sec:conclusion}
We present a communication chain construction algorithm for a
heterogeneous swarm of robots. Our approach is completely
decentralized: it requires only relative and local information from
neighbors. To tackle with robot failures, we exchange information and
bridge the chain as soon as it is broken. We assess the algorithm
performance with extensive simulations on up to 100 robots in five
different arenas. Real robot experiments with flying and ground robots
demonstrated the usability and robustness of the approach.




   



\bibliographystyle{IEEEtran}
\bibliography{icra20}

\end{document}

%% file: figures/chain_formation_illustaration_good.pdf_tex
\begingroup%
  \makeatletter%
  \providecommand\color[2][]{%
    \errmessage{(Inkscape) Color is used for the text in Inkscape, but the package 'color.sty' is not loaded}%
    \renewcommand\color[2][]{}%
  }%
  \providecommand\transparent[1]{%
    \errmessage{(Inkscape) Transparency is used (non-zero) for the text in Inkscape, but the package 'transparent.sty' is not loaded}%
    \renewcommand\transparent[1]{}%
  }%
  \providecommand\rotatebox[2]{#2}%
  \newcommand*\fsize{\dimexpr\f@size pt\relax}%
  \newcommand*\lineheight[1]{\fontsize{\fsize}{#1\fsize}\selectfont}%
  \ifx\svgwidth\undefined%
    \setlength{\unitlength}{513.53611392bp}%
    \ifx\svgscale\undefined%
      \relax%
    \else%
      \setlength{\unitlength}{\unitlength * \real{\svgscale}}%
    \fi%
  \else%
    \setlength{\unitlength}{\svgwidth}%
  \fi%
  \global\let\svgwidth\undefined%
  \global\let\svgscale\undefined%
  \makeatother%
  \begin{picture}(1,0.5236887)%
    \lineheight{1}%
    \setlength\tabcolsep{0pt}%
    \put(0,0){\includegraphics[width=\unitlength,page=1]{chain_formation_illustaration_good.pdf}}%
    \put(0.44939292,0.18938098){\color[rgb]{0,0,0}\makebox(0,0)[lt]{\begin{minipage}{0.21076884\unitlength}\raggedright \end{minipage}}}%
    \put(0.05142869,0.43327007){\color[rgb]{0,0,0}\makebox(0,0)[lt]{\lineheight{1.25000012}\smash{\begin{tabular}[t]{l}Worker\end{tabular}}}}%
    \put(0,0){\includegraphics[width=\unitlength,page=2]{chain_formation_illustaration_good.pdf}}%
    \put(0.05294082,0.38290736){\color[rgb]{0,0,0}\makebox(0,0)[lt]{\lineheight{1.25000012}\smash{\begin{tabular}[t]{l}Networker\end{tabular}}}}%
    \put(0,0){\includegraphics[width=\unitlength,page=3]{chain_formation_illustaration_good.pdf}}%
    \put(0.29489926,0.43494332){\color[rgb]{0,0,0}\makebox(0,0)[lt]{\lineheight{1.25000012}\smash{\begin{tabular}[t]{l}Target\end{tabular}}}}%
    \put(0.29762967,0.38366799){\color[rgb]{0,0,0}\makebox(0,0)[lt]{\lineheight{1.25000012}\smash{\begin{tabular}[t]{l}GS (Root)\end{tabular}}}}%
    \put(0,0){\includegraphics[width=\unitlength,page=4]{chain_formation_illustaration_good.pdf}}%
  \end{picture}%
\endgroup%

%% file: figures/movement.pdf_tex
\begingroup%
  \makeatletter%
  \providecommand\color[2][]{%
    \errmessage{(Inkscape) Color is used for the text in Inkscape, but the package 'color.sty' is not loaded}%
    \renewcommand\color[2][]{}%
  }%
  \providecommand\transparent[1]{%
    \errmessage{(Inkscape) Transparency is used (non-zero) for the text in Inkscape, but the package 'transparent.sty' is not loaded}%
    \renewcommand\transparent[1]{}%
  }%
  \providecommand\rotatebox[2]{#2}%
  \newcommand*\fsize{\dimexpr\f@size pt\relax}%
  \newcommand*\lineheight[1]{\fontsize{\fsize}{#1\fsize}\selectfont}%
  \ifx\svgwidth\undefined%
    \setlength{\unitlength}{283.56464216bp}%
    \ifx\svgscale\undefined%
      \relax%
    \else%
      \setlength{\unitlength}{\unitlength * \real{\svgscale}}%
    \fi%
  \else%
    \setlength{\unitlength}{\svgwidth}%
  \fi%
  \global\let\svgwidth\undefined%
  \global\let\svgscale\undefined%
  \makeatother%
  \begin{picture}(1,0.40490479)%
    \lineheight{1}%
    \setlength\tabcolsep{0pt}%
    \put(0,0){\includegraphics[width=\unitlength,page=1]{movement.pdf}}%
    \put(0.09620229,0.13876553){\color[rgb]{0,0,0}\makebox(0,0)[lt]{\lineheight{1.25}\smash{\begin{tabular}[t]{l}\small$x_i(t)$\end{tabular}}}}%
    \put(0.402827,0.13890049){\color[rgb]{0,0,0}\makebox(0,0)[lt]{\lineheight{1.25}\smash{\begin{tabular}[t]{l}\small$x_j(t)$\end{tabular}}}}%
    \put(0.23770919,0.13895593){\color[rgb]{0,0,0}\makebox(0,0)[lt]{\lineheight{1.25}\smash{\begin{tabular}[t]{l}\small$d_{ij}(t)$\end{tabular}}}}%
    \put(0.23845896,0.23731247){\color[rgb]{0,0,0}\makebox(0,0)[lt]{\lineheight{1.25}\smash{\begin{tabular}[t]{l}\small$d_{ij}(t+\Delta t)$\end{tabular}}}}%
    \put(0,0.20495568){\color[rgb]{0,0,0}\makebox(0,0)[lt]{\lineheight{1.25}\smash{\begin{tabular}[t]{l}\small$\Delta\overrightarrow{x_i}$\end{tabular}}}}%
    \put(0.46092801,0.20518848){\color[rgb]{0,0,0}\makebox(0,0)[lt]{\lineheight{1.25}\smash{\begin{tabular}[t]{l}\small$\Delta\overrightarrow{x_j}$\end{tabular}}}}%
    \put(0.02705392,0.2937136){\color[rgb]{0,0,0}\makebox(0,0)[lt]{\lineheight{1.25}\smash{\begin{tabular}[t]{l}\small$x_i(t+ \Delta t)$\end{tabular}}}}%
    \put(0.39266963,0.29761042){\color[rgb]{0,0,0}\transparent{0.99000001}\makebox(0,0)[lt]{\lineheight{1.25}\smash{\begin{tabular}[t]{l}\small$x_j(t+ \Delta t)$\end{tabular}}}}%
    \put(0,0){\includegraphics[width=\unitlength,page=2]{movement.pdf}}%
    \put(0.25309774,0.0229661){\color[rgb]{0,0,0}\makebox(0,0)[lt]{\begin{minipage}{0.07958799\unitlength}\raggedright (a)\end{minipage}}}%
    \put(0.64620694,0.02820206){\color[rgb]{0,0,0}\makebox(0,0)[lt]{\begin{minipage}{0.09945286\unitlength}\raggedright (b)\end{minipage}}}%
    \put(0,0){\includegraphics[width=\unitlength,page=3]{movement.pdf}}%
    \put(0.67125979,0.11226591){\color[rgb]{0,0,0}\makebox(0,0)[lt]{\lineheight{1.25}\smash{\begin{tabular}[t]{l}$x_i$(t-1)\end{tabular}}}}%
    \put(0.66851992,0.20900661){\color[rgb]{0,0,0}\makebox(0,0)[lt]{\lineheight{1.25}\smash{\begin{tabular}[t]{l}$x_j$(t-1)\end{tabular}}}}%
    \put(0.6820025,0.04847203){\color[rgb]{0,0,0}\makebox(0,0)[lt]{\lineheight{1.25}\smash{\begin{tabular}[t]{l} GS\end{tabular}}}}%
    \put(0,0){\includegraphics[width=\unitlength,page=4]{movement.pdf}}%
    \put(0.58316717,0.10949288){\color[rgb]{0,0,0}\makebox(0,0)[lt]{\lineheight{1.25}\smash{\begin{tabular}[t]{l}$d_{igs}$\end{tabular}}}}%
    \put(0.59644227,0.19993157){\color[rgb]{0,0,0}\makebox(0,0)[lt]{\lineheight{1.25}\smash{\begin{tabular}[t]{l}$d_{ij}$\end{tabular}}}}%
    \put(0.59388895,0.30322905){\color[rgb]{0,0,0}\makebox(0,0)[lt]{\lineheight{1.25}\smash{\begin{tabular}[t]{l}$d_{i\tau_{i}}$\end{tabular}}}}%
    \put(0.6788779,0.3668504){\color[rgb]{0,0,0}\makebox(0,0)[lt]{\lineheight{1.25}\smash{\begin{tabular}[t]{l}$\tau_{i}$\end{tabular}}}}%
    \put(0,0){\includegraphics[width=\unitlength,page=5]{movement.pdf}}%
    \put(0.88158425,0.0286522){\color[rgb]{0,0,0}\makebox(0,0)[lt]{\begin{minipage}{0.09945286\unitlength}\raggedright (c)\end{minipage}}}%
    \put(0,0){\includegraphics[width=\unitlength,page=6]{movement.pdf}}%
    \put(0.90753764,0.2069095){\color[rgb]{0,0,0}\makebox(0,0)[lt]{\lineheight{1.25}\smash{\begin{tabular}[t]{l}$x_i(t)$\end{tabular}}}}%
    \put(0.90755228,0.3077446){\color[rgb]{0,0,0}\makebox(0,0)[lt]{\lineheight{1.25}\smash{\begin{tabular}[t]{l}$x_j(t)$\end{tabular}}}}%
    \put(0.92306708,0.04847203){\color[rgb]{0,0,0}\makebox(0,0)[lt]{\lineheight{1.25}\smash{\begin{tabular}[t]{l} GS\end{tabular}}}}%
    \put(0,0){\includegraphics[width=\unitlength,page=7]{movement.pdf}}%
    \put(0.83253295,0.10820702){\color[rgb]{0,0,0}\makebox(0,0)[lt]{\lineheight{1.25}\smash{\begin{tabular}[t]{l}$d_{kgs}$\end{tabular}}}}%
    \put(0.8470638,0.29637115){\color[rgb]{0,0,0}\makebox(0,0)[lt]{\lineheight{1.25}\smash{\begin{tabular}[t]{l}$d_{ij}$\end{tabular}}}}%
    \put(0.91453654,0.3668504){\color[rgb]{0,0,0}\makebox(0,0)[lt]{\lineheight{1.25}\smash{\begin{tabular}[t]{l}$\tau_{i}$\end{tabular}}}}%
    \put(0,0){\includegraphics[width=\unitlength,page=8]{movement.pdf}}%
    \put(0.84493944,0.19607406){\color[rgb]{0,0,0}\makebox(0,0)[lt]{\lineheight{1.25}\smash{\begin{tabular}[t]{l}$d_{ik}$\end{tabular}}}}%
    \put(0.90975293,0.11152135){\color[rgb]{0,0,0}\makebox(0,0)[lt]{\lineheight{1.25}\smash{\begin{tabular}[t]{l}$x_k(t)$\end{tabular}}}}%
  \end{picture}%
\endgroup%

%% file: figures/method.tex
\tikzset{
  box1/.style={
           rectangle,
           draw=black, 
           thick,
           text centered},
    >=stealth',
    arrow1/.style={
           thick,
           shorten <=2pt,
           shorten >=2pt,
           }
}
\beginpgfgraphicnamed{method-sm}
\pgfdeclarelayer{background}
\pgfsetlayers{background,main}

    \begin{tikzpicture}
    
	\begin{axis}[
		width = 1.15\columnwidth,
		height = 5.5cm,
		xmin=0,
		xmax=100,
		ymin=0,
		ymax=100,	
		grid=none,
		axis line style={draw=none},
		tick style={draw=none},
		xticklabel=\empty,
		yticklabel=\empty,
		]


	\node 
		at (axis cs:0,0) 
		(fk1) {};

	\node 
		at (axis cs:30,50) 
		(fk3) {};

	\node[box1,
		dashed,
		rounded corners,
           	minimum height=1.5cm,
           	minimum width=4cm,
		text width=4cm,
		above right =-0.06cm and -0.0cm of fk1.north,
		] 
		 (cm) {};
	\node[below left] at ($(cm.north east)+(-0.0cm,0.0cm)$) {\footnotesize \texttt{Connectivity Module}};

	\node[box1,
		dashed,
		rounded corners,
           	minimum height=2.0cm,
           	minimum width=4cm,
		text width=4cm,
		above =0.15cm of cm,
		] 
		 (mm) {};
	\node[below left] at ($(mm.north east)+(-0.0cm,0.0cm)$) {\footnotesize \texttt{Mobility Module}};

		
	\node[box1,
		very thick,
		line width=0.1mm,
           	minimum height=0.5cm,
           	minimum width=1cm,
		text width=1cm,
		above right
		] 
		 (map) at ($(mm.north west)+(0.2cm,-1.35cm)$) {\footnotesize Map};

	\node[box1,
		very thick,
		line width=0.1mm,
           	minimum height=0.5cm,
           	minimum width=1.8cm,
		text width=1.8cm,
		above right
		] 
		 (pec) at ($(map.east)+(0.4cm,-0.2cm)$) {\footnotesize Path Finder Module};	

	\node[box1,
		very thick,
		line width=0.1mm,
           	minimum height=0.5cm,
           	minimum width=1.8cm,
		text width=1.8cm,
		below
		] 
		 (pp) at ($(pec.south)+(0.0cm,-0.0cm)$) {\footnotesize Path Planner};


	\node[box1,
		very thick,
		line width=0.1mm,
           	minimum height=0.5cm,
           	minimum width=1.5cm,
		text width=1.5cm,
		above right
		] 
		 (ni) at ($(cm.north west)+(0.2cm,-1.35cm)$) {\footnotesize Neighbor Information};

	\node[box1,
		very thick,
		line width=0.1mm,
           	minimum height=0.5cm,
           	minimum width=1.5cm,
		text width=1.5cm,
		above right
		] 
		 (cc) at ($(ni.east)+(0.3cm,-0.4cm)$) {\footnotesize Connectivity Estimator};


	\node[box1,
		very thick,
		line width=0.1mm,
           	minimum height=0.5cm,
           	minimum width=2cm,
		text width=2.1cm,
		right
		] 
		 (op) at ($(pec.east)+(1.3cm,0.757cm)$) {\footnotesize Operator};

	\node[box1,
		very thick,
		line width=0.1mm,
           	minimum height=0.5cm,
           	minimum width=2cm,
		text width=2.1cm,
		right
		] 
		 (sm) at ($(pec.east)+(1.3cm,-0.1cm)$) {\footnotesize Virtual Stigmergy};

	\node[box1,
		very thick,
		line width=0.1mm,
           	minimum height=0.5cm,
           	minimum width=2cm,
		text width=2cm,
		below
		] 
		 (cp) at ($(sm.south)+(0.0cm,-0.35cm)$) {\footnotesize Control Policy};

	\node[box1,
		very thick,
		line width=0.1mm,
           	minimum height=0.5cm,
           	minimum width=2cm,
		text width=2cm,
		below
		] 
		 (r) at ($(cp.south)+(0.0cm,-0.33cm)$) {\footnotesize Robot};

	\node[box1,
		very thick,
		line width=0.1mm,
           	minimum height=0.5cm,
           	minimum width=2cm,
		text width=2cm,
		below
		] 
		 (se) at ($(r.south)+(0.0cm,-0.33cm)$) {\footnotesize State estimation};



   \draw[->,rounded corners] (cc.east) -| node[above right = -0.5cm and -0.65cm]{($e_{p}$,$e_{c}$)} ++(0.8cm,0.5cm) |- (cp.west);
   \draw[->, rounded corners] (pp.east) -|  ++(0.8cm,0.0cm) |- node[above right = 0.15cm and -0.55cm] {($\pi$,PE,$\pi_{g}$)} (sm.west);
   \draw [->] (op.south) -- (sm.north)  node [midway, below right = -0.2cm and 0.1cm] {$\tau_i$};

   \draw [->] ($(sm.south)+(-0.5cm,-0.0cm)$) -- ($(cp.north)+(-0.5cm,-0.0cm)$)  node [midway, below right = -0.2cm and 0.0cm] { $\tau_i$};

   \draw [->] ($(sm.south)+(0.5cm,-0.0cm)$) -- ($(cp.north)+(0.5cm,-0.0cm)$)  node [midway, below right = -0.26cm and -0.0cm] { ($\pi$,PE,$\pi_{g}$)};

   \draw [->] (cp.south) -- (r.north)  node [midway, below right = -0.3cm and 0.0cm] { $u(t)$};

   \draw [->] (r.south) -- (se.north)  node [midway, below right = -0.3cm and -0.0cm] { $x(t)$};

   \draw[->, rounded corners] (se.east) --  ++(0.4cm,0.0cm) |- node[above left = -0.5cm and -0.1cm, rotate=90] {$\hat{x}(t)$} (cp.east);

	\end{axis}
    \end{tikzpicture}
    
\endpgfgraphicnamed

%% file: figures/interaction.tex
\tikzset{
  box1/.style={
           rectangle,
           draw=black, 
           thick,
           text centered},
   circle1/.style={
           circle,
	 	  draw,
 	      minimum width=0.9cm,
 	      minimum height=0.9cm,
          inner sep=0.000001cm,
          node distance = 1.3cm,
          text width=0.9cm,
          execute at begin node=\setlength{\baselineskip}{0.2cm},
           text centered}, 
    diamond1/.style={
    	draw, diamond, aspect=2,
    	inner sep=0.000001cm,
          node distance = 1.5cm,
          text width=0.9cm,
          execute at begin node=\setlength{\baselineskip}{0.2cm},
          text centered},   
    >=stealth',
    arrow1/.style={
           thick,
           shorten <=2pt,
           shorten >=2pt,
           }
}

\beginpgfgraphicnamed{interaction-sm}
\pgfdeclarelayer{background}
\pgfsetlayers{background,main}

    \begin{tikzpicture}
    
	\begin{axis}[
		width = 1.15\columnwidth,
		height = 6.12cm,
		xmin=0,
		xmax=100,
		ymin=0,
		ymax=100,	
		grid=none,
		axis line style={draw=none},
		tick style={draw=none},
		xticklabel=\empty,
		yticklabel=\empty,
		]


	\node 
		at (axis cs:9,95) 
		(fk1) {};

	
	\node[circle1,
		below right
		] 
		 (cp) at ($(fk1.south)+(0.0cm,-0.15cm)$) {\scriptsize Compute \\\& \\ Update Plan};


		
		
		

	\node[rotate=90, above left = 0.0cm and 0.25cm of cp ] (worker) {\scriptsize Worker}; 

    
	\node[diamond1, right = 0.2 cm of cp]
	(tr) {\scriptsize Target Reachable};

	\node[circle1, node distance = 1.5cm,
		right = 0.2cm of tr ] (sw) {\scriptsize Switch Worker};

	\node[diamond1,
		right = 0.2cm of sw ] (ew) {\scriptsize New Worker};

	\node[circle1,
		right = 0.2cm of ew ] (stf) {\scriptsize Free};
	
	\node[diamond1,node distance = 1.3cm, below right = 0.7cm and 0.918cm of cp]
	(Elect_k) {\scriptsize Elected $c_n$ Networkers};

	\node[circle1, node distance = 1.5cm,
		right = 0.3cm of Elect_k ] (ser) {\scriptsize Send request};

	\node[diamond1,
		right = -0.0cm and 0.3cm of ser ] (ane) {\scriptsize New\\ Edge?};

	\node[circle1,
		below  right = 0.7cm and 1.8cm of ser ] (ec) {\scriptsize  Extend Chain };
		
	\node[below right = 0.3cm and 0.3cm of ec, rotate=90 ] (net) {\scriptsize Networker};

	\node[diamond1,
		left =0.2cm of ec ] (net_emin) {\scriptsize  Networker required };
	
	\node[circle1,
		left = 0.2cm of net_emin ] (adj_pos) {\scriptsize Adjust Position Chain/ Target};

	\node[circle1,
		left = 0.2cm of adj_pos] (adj_rec) {\scriptsize Recover Edge};

	\node[diamond1,node distance = 1.5cm, left = 0.2cm of adj_rec]
	(C_broke) {\scriptsize Broken edges};



	\draw [->] ($(worker.south)+(-0.12cm,0.04cm)$) -- (cp.west); 

	\draw [->] (cp.east) -- (tr.west); 


	\draw [->] (tr.east) -- (sw.west) node [midway, above right = 0.1cm and -0.5cm] {\scriptsize False};

	\draw [->] (tr.south) -| node[above right = -0.2cm and 0.3cm]{\scriptsize True} ++(0.0cm,-0.2cm) -| (Elect_k.north);

	\draw [->] (sw.east) -- (ew.west); 

	\draw [->] (ew.east) -- (stf.west) node [midway, above right = 0.1cm and -0.5cm] {\scriptsize True}; 

\draw [->] (ew.north)  -| node[below left = -0.3cm and 0.2cm]{\scriptsize False} ++(0.0cm,0.32cm) -| (sw.north);

	\draw [->] (Elect_k.east) -- (ser.west) node [midway, above right = 0.1cm and -0.5cm] {\scriptsize False};

	\draw [->] (ser.east) -- (ane.west); 

	\draw [->] (ane.north) -| node[below left = -0.3cm and 0.0cm]{\scriptsize True} ++(0.0cm,0.24cm) -| (Elect_k.north); 


	\draw [->] (adj_pos.north west) to [out=160,in=20] (C_broke.north) node [midway, above right = 1.5cm and 1cm] {\scriptsize Fault Check}; 

	\draw [->] (C_broke.east) -- (adj_rec.west) node [midway, above = 0.2cm and 0.0cm] {\scriptsize True};

	\draw [->] (C_broke.south) -| node[above right = -0.3cm and 0.1cm]{\scriptsize False} ++(0.0cm,-0.3cm) -| (adj_pos.south west);

	\draw [->] (adj_rec.east) -- (adj_pos.west); 

	\draw [->] ($(net.north)+(0.1cm,0.00cm)$) -- (ec.east); 

	\draw [->] (ec.west) -- (net_emin.east); 
	
	\draw [->] (net_emin.west) -- (adj_pos.east) node [midway, above right = 0.1cm and -0.2cm] {\scriptsize False};


	\draw [->, dashed] (ane.east) -| node[above right = 0.0cm and 0.1cm]{\scriptsize True} ++(0.0cm,0.0cm) -| (ec.north east); 



	\draw [->] (adj_pos.south) -| ++(0.0cm,-0.15cm) -| (ec.south);

\draw [->] (Elect_k.south) -| node[above right = -0.2cm and 0.1cm]{\scriptsize True} ++(0.0cm,-0.1cm) -| (adj_pos.north);
 
 \draw [->, dashed] (net_emin.north) -| node[below right = -0.3cm and 0.1cm]{\scriptsize True} ++(0.0cm,0.1cm) -| (ser.south);
 
 \draw [->] (ane.south) -| node[below right = -0.1cm and 0.1cm]{\scriptsize False} ++(0.0cm,-0.2cm) -| (ser.south east);

	\end{axis}
    \end{tikzpicture}
    
\endpgfgraphicnamed

%% file: figures/failure_tol_illus.pdf_tex
\begingroup%
  \makeatletter%
  \providecommand\color[2][]{%
    \errmessage{(Inkscape) Color is used for the text in Inkscape, but the package 'color.sty' is not loaded}%
    \renewcommand\color[2][]{}%
  }%
  \providecommand\transparent[1]{%
    \errmessage{(Inkscape) Transparency is used (non-zero) for the text in Inkscape, but the package 'transparent.sty' is not loaded}%
    \renewcommand\transparent[1]{}%
  }%
  \providecommand\rotatebox[2]{#2}%
  \newcommand*\fsize{\dimexpr\f@size pt\relax}%
  \newcommand*\lineheight[1]{\fontsize{\fsize}{#1\fsize}\selectfont}%
  \ifx\svgwidth\undefined%
    \setlength{\unitlength}{338.99527631bp}%
    \ifx\svgscale\undefined%
      \relax%
    \else%
      \setlength{\unitlength}{\unitlength * \real{\svgscale}}%
    \fi%
  \else%
    \setlength{\unitlength}{\svgwidth}%
  \fi%
  \global\let\svgwidth\undefined%
  \global\let\svgscale\undefined%
  \makeatother%
  \begin{picture}(1,0.26700802)%
    \lineheight{1}%
    \setlength\tabcolsep{0pt}%
    \put(0,0){\includegraphics[width=\unitlength,page=1]{failure_tol_illus.pdf}}%
    \put(1.17463165,-0.5039953){\color[rgb]{0,0,0}\makebox(0,0)[lt]{\begin{minipage}{0.47698006\unitlength}\raggedright \end{minipage}}}%
    \put(0.82223347,-0.16422159){\color[rgb]{0,0,0}\makebox(0,0)[lt]{\begin{minipage}{0.11243226\unitlength}\raggedright \end{minipage}}}%
    \put(0.01215069,0.17098218){\color[rgb]{0,0,0}\makebox(0,0)[lt]{\lineheight{1.25}\smash{\begin{tabular}[t]{l}\normalsize$r_{p-2}$\end{tabular}}}}%
    \put(0.07099024,0.24020659){\color[rgb]{0,0,0}\makebox(0,0)[lt]{\lineheight{1.25}\smash{\begin{tabular}[t]{l} \normalsize $S_{p-1}$\end{tabular}}}}%
    \put(0.10739103,0.14539299){\color[rgb]{0,0,0}\makebox(0,0)[lt]{\lineheight{1.25}\smash{\begin{tabular}[t]{l}\normalsize $r_{p-1}$\end{tabular}}}}%
    \put(0.18248594,0.12102704){\color[rgb]{0,0,0}\makebox(0,0)[lt]{\lineheight{1.25}\smash{\begin{tabular}[t]{l}\normalsize $r_{p}$\end{tabular}}}}%
    \put(0,0){\includegraphics[width=\unitlength,page=2]{failure_tol_illus.pdf}}%
    \put(0.28342087,0.20964774){\color[rgb]{0,0,0}\makebox(0,0)[lt]{\lineheight{1.25}\smash{\begin{tabular}[t]{l}\normalsize $S_{p+1}$\end{tabular}}}}%
    \put(0.29530103,0.11529389){\color[rgb]{0,0,0}\makebox(0,0)[lt]{\lineheight{1.25}\smash{\begin{tabular}[t]{l}\normalsize $r_{p+1}$\end{tabular}}}}%
    \put(0,0){\includegraphics[width=\unitlength,page=3]{failure_tol_illus.pdf}}%
    \put(0.16887485,0.04433761){\color[rgb]{0,0,0}\makebox(0,0)[lt]{\lineheight{1.25}\smash{\begin{tabular}[t]{l}\normalsize $2*d_s$\end{tabular}}}}%
    \put(0.12734512,0.18632561){\color[rgb]{0,0,0}\makebox(0,0)[lt]{\lineheight{1.25}\smash{\begin{tabular}[t]{l}\normalsize $u_{p-1}^{pref}$\end{tabular}}}}%
    \put(0.21801026,0.15845031){\color[rgb]{0,0,0}\makebox(0,0)[lt]{\lineheight{1.25}\smash{\begin{tabular}[t]{l}\normalsize $u_{p+1}^{pref}$\end{tabular}}}}%
    \put(0,0){\includegraphics[width=\unitlength,page=4]{failure_tol_illus.pdf}}%
    \put(0.55293445,0.17487277){\color[rgb]{0,0,0}\makebox(0,0)[lt]{\lineheight{1.25}\smash{\begin{tabular}[t]{l}\normalsize $r_{p-2}$\end{tabular}}}}%
    \put(0.5114545,0.12643079){\color[rgb]{0,0,0}\makebox(0,0)[lt]{\lineheight{1.25}\smash{\begin{tabular}[t]{l} \normalsize $S_{p-1}$\end{tabular}}}}%
    \put(0.60058322,0.12277439){\color[rgb]{0,0,0}\makebox(0,0)[lt]{\lineheight{1.25}\smash{\begin{tabular}[t]{l}\normalsize $r_{p-1}$\end{tabular}}}}%
    \put(0.62555073,0.14257364){\color[rgb]{0,0,0}\makebox(0,0)[lt]{\lineheight{1.25}\smash{\begin{tabular}[t]{l}\normalsize $r_{p}$\end{tabular}}}}%
    \put(0,0){\includegraphics[width=\unitlength,page=5]{failure_tol_illus.pdf}}%
    \put(0.68757927,0.19315284){\color[rgb]{0,0,0}\makebox(0,0)[lt]{\lineheight{1.25}\smash{\begin{tabular}[t]{l}\normalsize $S_{p+1}$\end{tabular}}}}%
    \put(0.69449904,0.12297469){\color[rgb]{0,0,0}\makebox(0,0)[lt]{\lineheight{1.25}\smash{\begin{tabular}[t]{l}\normalsize $r_{p+1}$\end{tabular}}}}%
    \put(0,0){\includegraphics[width=\unitlength,page=6]{failure_tol_illus.pdf}}%
    \put(0.63578781,0.04015779){\color[rgb]{0,0,0}\makebox(0,0)[lt]{\lineheight{1.25}\smash{\begin{tabular}[t]{l}\normalsize $d_s$\end{tabular}}}}%
    \put(0,0){\includegraphics[width=\unitlength,page=7]{failure_tol_illus.pdf}}%
    \put(0.59280346,0.2434779){\color[rgb]{0,0,0}\makebox(0,0)[lt]{\lineheight{1.25}\smash{\begin{tabular}[t]{l}\normalsize $S_{p-1} \cup S_{p+1} \neq \emptyset$\end{tabular}}}}%
    \put(0,0){\includegraphics[width=\unitlength,page=8]{failure_tol_illus.pdf}}%
    \put(0.37671803,0.13659542){\color[rgb]{0,0,0}\makebox(0,0)[lt]{\lineheight{1.25}\smash{\begin{tabular}[t]{l}\normalsize $\pi(r_n)$\end{tabular}}}}%
    \put(0.83074721,0.15405808){\color[rgb]{0,0,0}\makebox(0,0)[lt]{\lineheight{1.25}\smash{\begin{tabular}[t]{l}\normalsize $\pi(r_n)$\end{tabular}}}}%
    \put(0.14638331,0.00311122){\color[rgb]{0,0,0}\makebox(0,0)[lt]{\lineheight{1.25}\smash{\begin{tabular}[t]{l}(a)\end{tabular}}}}%
    \put(0.64417786,0.00358529){\color[rgb]{0,0,0}\makebox(0,0)[lt]{\lineheight{1.25}\smash{\begin{tabular}[t]{l}(b)\end{tabular}}}}%
  \end{picture}%
\endgroup%